\documentclass{article} 
\usepackage{iclr2021_conference,times}
\usepackage{graphicx}  

\usepackage{amsmath,amsfonts,bm}

\def\eqref#1{equation~\ref{#1}}

\def\1{\bm{1}}

\DeclareMathAlphabet{\mathsfit}{\encodingdefault}{\sfdefault}{m}{sl}
\SetMathAlphabet{\mathsfit}{bold}{\encodingdefault}{\sfdefault}{bx}{n}

\usepackage{tabularx}

\usepackage{subfig}
\usepackage{hyperref}
\usepackage{url}

\title{Improving ResNet-9 Generalization Trained on Small Datasets}

 \author{\textbf{Omar Mohamed Awad}\thanks{Corresponding author} , \textbf{Habib Hajimolahoseini}\thanks{Same contribution as the first author in writing the paper} , \textbf{Michael Lim}, \textbf{Gurpreet Gosal}, \\
 \textbf{Walid Ahmed}, \textbf{Yang Liu}, \textbf{Gordon Deng} \\
 Ascend Team, Toronto Research Center, Huawei Technologies\\
 \texttt{omar.mohamedawad@huawei.com}\\
 \texttt{habib.hajimolahoseini@huawei.com}
 }

\iclrfinalcopy 
\begin{document}

\maketitle

\begin{abstract}
This paper presents our proposed approach that won the first prize at the ICLR competition "Hardware Aware Efficient Training".
The challenge is to achieve the highest possible accuracy in
an image classification task in less than 10 minutes. 
The training is done on a small dataset of 5000 images picked randomly from CIFAR-10 dataset. 
The evaluation is performed by the competition organizers on a secret dataset with 1000 images of the same size. 
Our approach includes applying a series of technique for improving the generalization of ResNet-9 including: sharpness aware optimization, label smoothing, gradient centralization, input patch whitening as well as meta-learning based training.
Our experiments show that the ResNet-9 can achieve the accuracy of 88\% while trained only on a 10\% subset of CIFAR-10 dataset in less than 10 minuets. 

\end{abstract}

\section{Introduction}
Deep learning models have become larger and larger over the past few years which leads to models with huge numbers of training parameters. 
Training such a huge model requires a large amount of memory and computational power \cite{hajimolahoseinistrategies, hajimolahoseinicompressing,walid2022training,walid2022speeding} as well as big datasets. 
However, for some applications, especially on the edge devices with online learning capabilities, the memory and training time could be limited and a large dataset may not be available too \cite{li2021short, hajimolahoseini2023methods, hajimolahoseini2019deep}.
In cases like this, the model is highly prone to the issue of overfitting and may also not be able to generalize well. 
Hence, having a model which is able to learn fast on a small dataset and generalize well is highly beneficial \cite{hajimolahoseini2018ecg}. 

There are some techniques proposed in the literature which try to improve the training efficiency using different approaches during training  \cite{yong2020gradient, foret2020sharpness, zhang2018shufflenet}.
For a detailed review of related work the reader is referred to \cite{coleman2017dawnbench}.
In this competition, the goal is to reach to the highest possible accuracy in an image classification task in less than 10 minuets.
The training is performed on a 10\% subset of CIFAR10 dataset including 10 classes, each with 500 training samples and 100 test samples
of $32\times32$ pixels RGB images \cite{krizhevsky2009learning}. 
The evaluation
is performed on mini-ImageNet dataset (hidden
at development time) of size similar to the development
dataset \cite{deng2009imagenet}. 
No pre-trained weights are allowed so the models are trained from scratch. 
More details about the proposed methodology is presented in the following sections. 

\section{Methodology}
\subsection{Baseline model}
Inspired by most of the recent approaches in the similar competitions \cite{coleman2017dawnbench}, we adopt the well-known ResNet-9 architecture \cite{he2016deep} as our baseline. 
Due to its relatively small size, it could be beneficial in preventing the over-fitting issue.
However, for training in such a short time on a small dataset, some strategies need to be applied in order to increase the accuracy of the model and its generalization ability.


\subsection{Optimizer}
During the training of the baseline model, we observed that the generalization of first-order optimizers such as SGD was sub-par, since the test accuracy of ResNet-9 trained on the 10\% of CIFAR10 dataet could not reach higher than 76.65\%.
Therefore, a Sharpness Aware Minimization (SAM) technique is employed to improve the generalization of the model \cite{foret2020sharpness}.

For models with a very large capacity, the value of training loss does not necessarily guarantee the ability of the model to generalize well \cite{foret2020sharpness}. 
In contrast, SAM optimizer minimizes both the value and sharpness of the loss function at the same time. 
In other words, it looks for parameters in the surrounding area that have a uniformly low loss.
This method converts the minimization problem into a min-max optimization on which SGD could be performed more efficiently.
Instead of looking for parameter values that show a low training loss, SAM searches for those parameter values whose entire surrounding area has a uniformly low training loss.
It means that the neighborhoods that have both low loss and low curvature \cite{foret2020sharpness}. 

\subsection{Gradient Centralization}
Another technique that can help improving the generalization of small models is Gradient Centralization (GC) \cite{yong2020gradient}.
It centralizes the gradients so that they have zero mean. 
This is done by creating a projected gradient descent method with a constrained loss function. 
By regularizing the solution space of model parameters, GC helps to reduce the possibility of overfitting on training data and improving generalization of trained models, especially for small datasets.


\subsection{Improved Preprocessing}
We use a set of techniques called Improved Preprocessing (IP). The methods are described in the following sections:
\subsubsection{Label Smoothing} 
In order to improve our training strategy, we use label smoothing technique \cite{pereyra2017regularizing}. 
It blends the one-hot target probabilities with a uniform distribution over class labels inside the cross entropy loss as follow:
\begin{equation}\label{eq:labelsmoothing}
y_{ls} = (1-\alpha) \times y_{hot} + \frac{\alpha}{K}
\end{equation}
where $y_{ls}$ and $y_{hot}$ represent the label smoothed and one-hot probabilities and $K$ and $\alpha$ are the number of classes ($K=10$ for our case) and smoothing factor, respectively. 
In our experiments, we set the smoothing factor $\alpha=0.1$ and number of classes are $K=10$. 
This helps to stabilize the output distribution and prevents the network from making overconfident predictions which might inhibit further training.

\subsubsection{Weight Decay} 
Weight decay is another regularization method that we use which keeps the model weights as small as possible:
\begin{equation}\label{eq:weight_decay}
L_{new}(w) = L_{org}(w) + \lambda{w^{T}w}
\end{equation}
where $L_{new}(w)$ and $L_{org}(w)$ are the new and original loss functions and $\lambda$ is the decay factor which we set to $\lambda = 0.0005$.. 
The small values of weights guarantees that the network behaviour will not change much if we change a few random inputs which in turn makes it difficult for the regularized network to learn local noise in the data.

\subsubsection{Activation Function} 
We replaced the RELU activation function of ResNet-9 with CELU \cite{barron2017continuously} which helps improving the generalization since smoothed functions lead to a less expressive function class:
\begin{equation}\label{eq:celu}
\begin{split}
\text{CELU}(x,\alpha) = \begin{Bmatrix} x & x \geq 0 \\
 \alpha (\exp(\frac{x}{\alpha})-1) & \text{otherwise} \end{Bmatrix}\end{split}
\end{equation}
According to \ref{eq:celu}, CELU converges to ReLU as $\alpha \to 0$.
In our experiments we set $\alpha = 0.3$.


\subsubsection{Input Patch Whitening}
In order to reduce the covariance between channels and pixels we apply PCA whitening to $3{\times}3$ patches of inputs as an initial $3{\times}3$ convolution with fixed (non-trainable) weights followed by a trainable $1{\times}1$ convolution \cite{jegou2012negative}. 

\subsection{Meta-learning based Training}
As a mechanism for generalizing the knowledge learned over many few-shot learning tasks, meta-learning is a promising training approach for few-shot learning problems.
A collection of different (but similar) few-shot learning tasks are learned in parallel to learn representations that are common to all tasks (e.g. Omniglot dataset). 

\begin{figure}[h]
\begin{center}
\includegraphics[width=\textwidth]{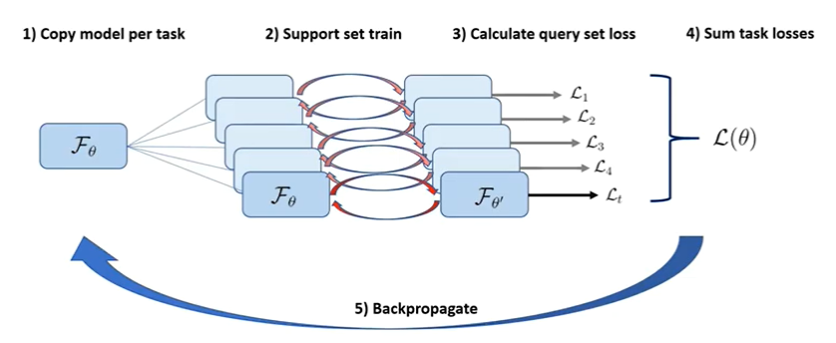}
\end{center}
\caption{Meta-learning Algorithm}
\label{fig:metalearning1}
\end{figure}

This meta-learning approach has been reframed as a single-task algorithm for training on small dataset (10 classes of Mini-ImageNet) – named Meta-Learning based Training Procedure (MLTP). 
MLTP was applied in hopes to get good generalization from a small dataset, and ultimately higher classification accuracy.
In this approach, the Mini-ImageNet dataset is broken up into 2 tasks (each containing 250 samples of each class), and applied as a 2-task MLTP problem. The base architecture used is ResNet9.
This process is depicted in Fig.\ref{fig:metalearning1}.

\section{Evaluation}
The hardware used in this competition is an Nvidia V100 GPU which has a memory of 32GB, running with an Intel(R) Xeon(R) Gold 6230 CPU @ 2.10GHz processor, with a RAM of 12GB.
We randomly pick 5000 images from the CIFAR-10 dataset (500 images from each of the 10 classes) for training. 

The experimental results are shown in Table \ref{tbl:table}. According to this table, although each of the SAM, GC and meta-learning techniques can improve the accuracy of the baseline model individually, however,
combining the SAM optimizer and the IP techniques (SAM+IP) can boost the model accuracy by 9.95\% and 7.7\% on Mini-CIFAR10 and Mini-ImageNet datasets, respectively when comparing to the baseline model. 

 \begin{table}[htb]
		\centering\footnotesize
		\caption{Accuracy \& Performance Results}
		\label{tbl:table}
		 \begin{tabular}{|c|c|c|c|c|}
         \hline
         \textbf{Model} & \multicolumn{2}{|c|}{\textbf{Test Accuracy (\%)}} & \multicolumn{2}{|c|}{\textbf{Epochs (total runtime within 10 minutes)}} \\ \hline
         & Mini-CIFAR10 & Mini-ImageNet & Mini-CIFAR10 & Mini-ImageNet \\ \hline
         Baseline & 76.65 & 80.3 &  168 & 159 \\ \hline
         SAM & 79.9 & 83.25 & 151 &  161 \\ \hline
         \textbf{SAM+IP} & \textbf{86.6} & \textbf{88} & \textbf{ 200} & \textbf{200} \\ \hline
         SAM+GC & 83.15 & 84.2 &  99 &  102 \\ \hline
         Meta-learning & 82.87 & 85.04 &  132 & 136 \\ \hline
		\end{tabular}
\end{table}

\section{Conclusion}
In this paper, we adopted some techniques for improving the generalization of ResNet-9 model when training on a small dataset in a very short time. 
Experimental results reveal that thechniques such as SAM, GC, IP and meta learning can boost the performance of the baseline model especially when they are combined together. 
It is also worth mentioning that the proposed methods are orthogonal to each other and therefore they can be combined together. 
Due to time limitation of the competition, we did not experiment with all different combinations of the proposed methods. This is left for future work.



\bibliography{iclr2021_conference}
\bibliographystyle{iclr2021_conference}


\end{document}